\documentclass[10pt]{article} 
\usepackage[accepted]{tmlr}

\usepackage{amsmath,amsfonts,bm}

\def\eqref#1{equation~\ref{#1}}

\def\1{\bm{1}}

\DeclareMathAlphabet{\mathsfit}{\encodingdefault}{\sfdefault}{m}{sl}
\SetMathAlphabet{\mathsfit}{bold}{\encodingdefault}{\sfdefault}{bx}{n}

\usepackage[colorlinks=true,linkcolor=teal,citecolor=teal,urlcolor=magenta]{hyperref}
\usepackage{url}
\usepackage{booktabs}
\usepackage{amsfonts}
\usepackage{nicefrac}
\usepackage{microtype}
\usepackage{xcolor}
\usepackage{amsmath}
\usepackage{amssymb}
\usepackage{mathtools}
\usepackage{amsthm}
\usepackage{svg}
\usepackage{multirow}
\usepackage{siunitx}
\usepackage{placeins}
\usepackage{caption}
\usepackage{subcaption}
\usepackage{tabularx}
\usepackage{wrapfig}
\usepackage{twemojis}
\usepackage{diagbox}

\title{V{\Large 1}T: large-scale mouse V1 response prediction\\using a Vision Transformer}

\author{
    \name \hspace{-4pt}Bryan M. Li$^1$ \email bryan.li@ed.ac.uk
    \AND
    \name Isabel M. Cornacchia$^1$ \email isabel.cornacchia@ed.ac.uk
    \AND
    \name Nathalie L. Rochefort$^{2, 3}$ \email n.rochefort@ed.ac.uk
    \AND
    \name Arno Onken$^1$ \email aonken@ed.ac.uk
    \AND
    \addr $^1$School of Informatics, University of Edinburgh\\
    \addr $^2$Centre for Discovery Brain Sciences, University of Edinburgh\\
    \addr $^3$Simons Initiative for the Developing Brain, University of Edinburgh
}

\def\month{08}  
\def\year{2023} 
\def\openreview{\url{https://openreview.net/forum?id=qHZs2p4ZD4}} 

\begin{document}

\maketitle

\begin{abstract}
Accurate predictive models of the visual cortex neural response to natural visual stimuli remain a challenge in computational neuroscience. In this work, we introduce V{\small 1}T, a novel Vision Transformer based architecture that learns a shared visual and behavioral representation across animals. We evaluate our model on two large datasets recorded from mouse primary visual cortex and outperform previous convolution-based models by more than 12.7\% in prediction performance. Moreover, we show that the self-attention weights learned by the Transformer correlate with the population receptive fields. Our model thus sets a new benchmark for neural response prediction and can be used jointly with behavioral and neural recordings to reveal meaningful characteristic features of the visual cortex. Code available at \href{https://github.com/bryanlimy/V1T}{github.com/bryanlimy/V{\footnotesize 1}T}.
\end{abstract}

\section{Introduction} \label{introduction}

Understanding how the visual system processes information is a fundamental challenge in neuroscience. Predictive models of neural responses to naturally occurring stimuli have shown to be a successful approach toward this goal, serving the dual purpose of generating new hypotheses about biological vision~\citep{bashivan2019neural, walker2019inception, ponce2019evolving} and bridging the gap between biological and computer vision~\citep{li2019learning, sinz2019engineering, safarani2021towards}. This approach relies on the idea that high performing predictive models, which explain a large part of the stimulus-driven variability, have to account for the nonlinear response properties of the neural activity, thus allowing for the identification of the underlying computations of the visual system~\citep{carandini2005we}.

An extensive amount of work on the primary visual cortex (V1) has been dedicated to building quantitative models that accurately describe neural responses to visual stimuli, starting from simple linear-nonlinear models~\citep{heeger1992half, jones1987two}, energy models~\citep{adelson1985spatiotemporal} and multi-layer models~\citep{lehky1992predicting, lau2002computational, prenger2004nonlinear}. These models, based on neurophysiological data, provide a powerful framework to test hypotheses about neural functions and investigate the principles of visual processing. With the increased popularity of deep neural networks (DNNs) in computational neuroscience in recent years~\citep{kietzmann2018deep, richards2019deep, li2020calciumgan, li2021neuronal}, DNNs have set new standards of prediction performance~\citep{antolik2016model, klindt2017neural, ecker2018rotation, zhang2019convolutional}, allowing for a more extensive exploration of the underlying computations in sensory processing~\citep{walker2019inception, bashivan2019neural, burg2021learning}. 

DNN-based models are characterized by two main approaches. On the one hand, task-driven models rely on pre-trained networks optimized on standard vision tasks, such as object recognition, in combination with a readout mechanism to predict neural responses~\citep{yamins2014performance, cadieu2014deep, cadena2019deep}. With the goal of explaining the evolutionary and developmental constraints of the visual system, task-driven models have proven to be successful for predicting visual responses in primates~\citep{yamins2016using, cadena2019deep} and mice~\citep{nayebi2022mouse} by obtaining a shared generalized representation of the visual input across animals. On the other hand, data-driven models aim to build a predictive model on large-scale datasets without any assumption on the functional properties of the network. These models share a common representation by being trained end-to-end directly on data from thousands of neurons, and they have been shown to be successful as predictive models for the mouse visual cortex~\citep{lurz2021generalization, franke2022state}. This approach allows us to identify core components that can be insightful when studying nontrivial computational properties of cortical neurons, especially in combination with experimental verification~\citep{walker2019inception}.

Data-driven models for prediction of visual responses across multiple animals typically employ the core-readout framework~\citep{klindt2017neural, cadena2019deep, lurz2021generalization, burg2021learning, franke2022state}. Namely, a core module which learns a shared latent representation of the visual stimuli across the animals, followed by animal-specific linear readout modules to predict neural responses given the latent features. This architecture enforces the nonlinear computations to be performed by the shared core, which can in principle capture general characteristic features of the visual cortex~\citep{lurz2021generalization}. The readout models then learn the animal-specific mapping from the shared representation of the input to the individual neural responses. With the advent of large-scale neural recordings, datasets that consist of thousands or even hundreds of thousands of neurons are becoming readily available~\citep{stosiek2003vivo, steinmetz2021neuropixels}. This has led to an increase in the number of parameters needed in the readout network to account for the large number of neurons, hence significant effort in neural predictive modeling has been dedicated to develop more efficient readout networks. On the other hand, due to their effectiveness and computation efficiency~\citep{goodfellow2016deep}, convolutional neural networks (CNNs) are usually chosen as the shared representation model.

Recently, Vision Transformer (ViT, \citealt{dosovitskiy2021an}) has achieved excellent results in a broad range of computer vision tasks~\citep{han2022survey} and Transformer-based~\citep{vaswani2017attention} models have become increasingly popular in computational neuroscience~\citep{tuli2021convolutional, schneider2022learnable, whittington2022relating}. For instance, \citet{ye2021ndt} proposed a Neural Data Transformer to model spike trains, which was extended by \citet{le2022stndt} using a Spatial Transformer to achieve state-of-the-art performance in 4 neural datasets. \citet{berrios2022joint} introduced a data augmentation and adversarial training procedure to train a dual-stream Transformer which showed strong performance in predicting monkey V4 responses. In modeling the mouse visual cortex, \citet{conwell2021neural} experimented with a wide range of out-of-the-box DNNs, including CNNs and ViTs, to compare their representational similarity when pre-trained versus randomly initialized. Here, we explore the benefits of the ViT convolution-free approach and self-attention mechanism as the core representation learner in a data-driven neural predictive model. Note that, in this text, the term ``attention'' strictly refers to the self-attention layer in Transformers~\citep{vaswani2017attention}, which is distinct from the perceptual process of ``attention'' in the neuroscience literature.

Since neural variability shows a significant correlation with the internal brain state~\citep{pakan2016behavioral, pakan2018action, stringer2019spontaneous}, information about behavior can greatly improve visual system models in the prediction of neural responses~\citep{bashiri2021flow, franke2022state}. To exploit this relationship, we also investigate a principled mechanism in the model architecture to integrate behavioral states with visual information.

Altogether, we propose V{\small 1}T, a novel ViT-based architecture that can capture visual and behavioral representations of the mouse visual cortex. This core architecture, in combination with an efficient per-animal readout~\citep{lurz2021generalization}, outperforms the previous state-of-the-art model by 12.7\% and 19.1\% on two large-scale mouse V1 datasets~\citep{willeke2022sensorium, franke2022state}, which consist of neural recordings of thousands of neurons across over a dozen behaving rodents in response to thousands of natural images. Moreover, we show that the attention weights learned by the core module correlate with behavioral variables, such as pupil direction. This link between the model and the visual cortex activity is useful for pinpointing how behavioral variables affect neural activity.

\section{Neural data} \label{neural-data}

We considered two large-scale neural datasets for this work, \textsc{Dataset S}\footnote{The \href{https://sensorium2022.net/}{Sensorium Challenge} held at \href{https://neurips.cc/Conferences/2022/CompetitionTrack}{NeurIPS 2022 Competition Track Program}} by \citet{willeke2022sensorium} and \textsc{Dataset F} by \citet{franke2022state}. These two datasets consist of V1 recordings from behaving rodents in response to thousands of natural images, providing an excellent platform to evaluate our proposed method and compare it against previous visual predictive models. 

We first briefly describe the animal experiment in \textsc{Dataset S}. A head-fixed mouse was placed on a cylindrical treadmill with a \SI{25}{inch} monitor placed \SI{15}{\cm} away from the animal's left eye and more than 7,000 neurons from layer L2/3 in V1 were recorded via two-photon calcium imaging. Note that the position of the monitor was selected such that the stimuli were shown to the center of the recorded population receptive field. Gray-scale images $x_\text{image} \in \mathbb{R}^{c=1 \times h \times w}$ from ImageNet~\citep{deng2009imagenet} were presented to the animal for \SI{500}{\ms} with a blank screen period of \SI{300} to \SI{500}{\ms} between each presentation. Neural activities were accumulated between \SI{50} and \SI{500}{\ms} after each stimulus onset. In other words, for a given neuron $i$ in trial (stimulus presentation) $t$, the neural response is represented by a single value $r_{i, t}$. In addition, the anatomical coordinates of each neuron as well as four behavioral variables $x_\text{behaviors}$ were recorded alongside with the calcium responses. These variables include pupil dilation, the derivative of the pupil dilation, pupil center (2d-coordinates) and running speed of the animal. Each recording session consists of up to 6,000 image presentations (i.e. trials), where 5,000 unique images are combined with 10 repetitions of 100 additional unique images, randomly intermixed. The 1,000 trials with repeated images are used as the test set and the rest are divided into train and validation sets with a split ratio of 90\% and 10\% respectively. In total, data from 5 rodents\footnote{2 additional mice were used in the Sensorium challenge~\citep{willeke2022sensorium} and their test sets are not publicly available.} (\textsc{Mouse A} to \textsc{E}) were recorded in this dataset.

\textsc{Dataset F} follows largely the same experimental setup with the following distinction: colored images (UV-colored and green-colored, i.e. $x_\text{image} \in \mathbb{R}^{c=2 \times h \times w}$) from ImageNet were presented on a screen placed \SI{12}{\cm} away from the animal; 4,500 unique colored and 750 monochromatic images were used as the training set and an additional 100 unique colored and 50 monochromatic images were repeated 10 times throughout the recording; in total, 10 rodents (\textsc{Mouse F} to {O}) were used in the experiment with $1,000$ V1 neurons recorded from each animal. Table~\ref{table:recording-info} summarizes the experimental information from both datasets.

\section{Previous work} \label{previous-work}

A substantial body of work has recently focused on predictive models of cortical activity that learn a shared representation across neurons~\citep{klindt2017neural, cadena2019deep, lurz2021generalization, burg2021learning, franke2022state}, which stems from the idea in systems neuroscience that cortical computations share common features across animals~\citep{olshausen1996emergence}. In DNN models, these generalizing features are learned in a nonlinear core module, then a subsequent neuron-specific readout module linearly combines the relevant features in this representation to predict the neural responses. Recently, \citet{lurz2021generalization} and \citet{franke2022state} introduced a shared CNN core and animal-specific Gaussian readout combination that achieved excellent performance in mouse V1 neural response prediction, and this is the current state-of-the-art model on large-scale benchmarks including \textsc{Dataset S} and \textsc{Dataset F}. Here, we provide a brief description for each of the modules in their proposed architecture, which our work is built upon.

\textbf{CNN core}. Typically, the core module learns the shared visual representation via a series of convolutional blocks~\citep{cadena2019deep, lurz2021generalization, franke2022state}. In \citet{lurz2021generalization}, given an input image $x_\text{image} \in \mathbb{R}^{c \times h \times w}$, the CNN core with filter size $k$ outputs a latent representation vector $z \in \mathbb{R}^{d \times h' \times w'}$ where $h' = h - k + 1$, $w' = w - k + 1$ and $d$ is the hidden dimension. The CNN core, after an exhaustive Bayesian hyperparameter search to optimize for the validation performance, has an output dimension of $z \in \mathbb{R}^{d \times h'=28 \times w'=56}$. Previous works have shown correlation between behaviors and neural variability, and that the behavioral variables can significantly improve neural predictivity~\citep{niell2010modulation, reimer2014pupil, stringer2019spontaneous, bashiri2021flow}. To that end, \citet{franke2022state} proposed to integrate the behavioral variables $x_\text{behaviors} \in \mathbb{R}^v$ with the visual stimulus by duplicating each variable to a $h \times w$ matrix and concatenating them with $x_\text{image}$ in the channel dimension, resulting in an input vector of $\mathbb{R}^{(c + v) \times h \times w}$.

\textbf{Readout}. To compute the neural response of neuron $i$ from mouse $m$ with $n_m$ neurons, the readout module $\texttt{R}_m: \mathbb{R}^{d \times h' \times w'} \rightarrow \mathbb{R}^{n_m}$ by \citet{lurz2021generalization} computes a linear regression of the core representation $z$ with weights $w_i \in \mathbb{R}^{w' \times h' \times c}$, followed by an ELU activation with an offset of 1 (i.e. $o = \text{ELU}(\texttt{R}_m(z)) + 1$), which keeps the response positive. The regression is performed by a Gaussian readout, which learns the parameters of a 2d Gaussian distribution whose mean $\mu_i$ represents the center of the receptive field of the neuron in the image space and whose variance quantifies the uncertainty of the receptive field position, which decreases over training. The response is thus obtained as a linear combination of the feature vector of the core at a single spatial position, which allows the model to greatly reduce the number of parameters per neuron in the readout. Notably, to learn the position $\mu_i$, the model also exploits the retinotopic organization of V1 by coupling the recorded cortical 2d coordinates of each neuron with the estimated center of the receptive field from the readout. Moreover, a shifter module is introduced to adjust (or shift) the $\mu_i$ receptive field center of neuron $i$ to account for the trial-to-trial variability due to eye movement~\citep{franke2022state}. The shifter network $\mathbb{R}^2 \rightarrow \mathbb{R}^2$ consists of 3 dense layers with hidden size of 5 and $\tanh$ activation; it takes as input the 2d pupil center coordinates and learns the vertical and horizontal adjustments needed to shift $\mu_i$.

\section{Methods} \label{methods}

The aim of this work is to design a neural predictive model $F(x_\text{image}, x_\text{behaviors})$ that can effectively incorporate both visual stimuli and behavioral variables to predict responses $o$ that are faithful to real recordings $r$ from mouse V1. With that goal, we first detail the core architectures proposed in this work, followed by the training procedure and evaluation metrics. Code used in this work is attached as supplementary material and will be made publicly available upon publication.

\subsection{V{\small 1}T core} \label{methods:model}

\begin{wrapfigure}{r}{0.5\textwidth}
    \begin{center}
        \includegraphics[width=0.4\textwidth]{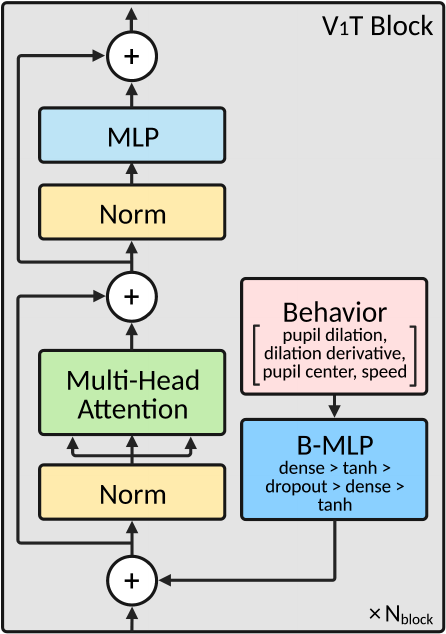}
    \end{center}
    \caption{Illustration of the V{\footnotesize 1}T block architecture.}
    \label{figure:v1t-architecture}
\end{wrapfigure}

Vision Transformers~\citep{dosovitskiy2021an}, or ViTs, have achieved competitive performance in many computer vision tasks, including object detection and semantic segmentation, to name a few~\citep{chen2020generative, carion2020end, strudel2021segmenter}. Here, we propose a data-driven ViT core capable of learning a shared representation of the visual stimuli that is relevant for the prediction of neural responses in the visual cortex. Moreover, we introduce an alternative approach in V{\small 1}T to encode behavioral variables in a more principled way when compared to previous methods and further improve the neural predictive performance of the overall model. 

The original ViT classifier is comprised of 3 main components: (1)~a tokenizer first encodes the 3d image (including channel dimension) into 2d patch embeddings, (2)~the embeddings are then passed through a series of Transformer~\citep{vaswani2017attention} encoder blocks, each consisting of a Multi-Head Attention (\texttt{MHA}) and a Multi-Layer Perceptron (\texttt{MLP}) module which requires 2d inputs, and finally (3) a classification layer outputs the class prediction. The following sections detail the modifications made to convert the vanilla ViT to a shared visual representation learner for the downstream readout modules. We additionally experiment with a number of recently proposed efficient ViTs that have been emphasized for learning from small to medium size datasets. 

\textbf{Tokenizer}. The tokenizer, or patch encoder, extracts non-overlapping squared patches of size $p \times p$ from the 2d image and projects each patch to embeddings $z_0$ of size $d$, i.e. $\mathbb{R}^{c \times h \times w} \rightarrow \mathbb{R}^{l \times (cp^2)} \rightarrow \mathbb{R}^{l \times d}$, where $l = hw/p^2$ is the number of patches. \citet{dosovitskiy2021an} proposed two tokenization methods in the original ViT, where patches can be extracted either (1) via a $p \times p$ sliding window over the height and width dimensions of the image, followed by a linear layer with $d$ hidden units, or (2) via a 2d convolutional layer with kernel size $p$ and $d$ filters. 

Transformer-based models benefit from (or even necessitate) pre-training on large datasets, in the magnitude of millions or even billions of samples, in order to obtain optimal performance~\citep{han2022survey}. In contrast, typical neural recordings in animal experiments are considerably smaller. To stay consistent with previous work, we instead focus on developing a core architecture that can be effectively trained on limited amount of data from scratch. To that end, we considered two recently introduced efficient ViT methods that are highly competitive in scarce data settings. \citet{lee2021vision} proposed Shifted Patch Tokenization (SPT) to combat the low inductive bias in ViTs and enable better learning from limited data. Conceptually, SPT allows additional (adjacent) pixel values to be included in each patch, thus improving the locality, or receptive field, of the model. Input image $x_\text{image} \in \mathbb{R}^{1 \times h \times w}$ is shifted spatially by $p / 2$ in one of the four diagonal directions (top-left, top-right, bottom-left, bottom-right) with zero padding and the four shifted images (i.e. each shifted in one diagonal direction) are then concatenated with the original image, resulting in a vector $\mathbb{R}^{5 \times h \times w}$, which can be processed by the two patch extraction approaches mentioned above. With a similar goal in mind, the Compact Convolutional Transformer (CCT, \citealt{hassani2021escaping}) was proposed as a convolutional tokenizer to learn the patch embeddings that can take advantage of the translation equivariance and locality inherent in CNNs. The proposed mini-CNN is fairly simple: it consists of a 2d convolution layer with a $p \times p$ kernel and filter size $d$, followed by ReLU activation and a max pool layer. In this work, we experimented with and compared all four tokenization methods: sliding window, a single 2d convolutional layer, SPT and CCT.

As ViTs are agnostic to the spatial structure of the data, a positional embedding is added to each patch to encode the relative position of the patches with respect to each other~\citep{dosovitskiy2021an, han2022survey} and this positional embedding can either be learned or sinusoidal. Finally, a learnable BERT~\citep{devlin2018bert} \texttt{[cls]} token is typically added to the patch embeddings (i.e. $z_0 \in \mathbb{R}^{(l + 1) \times d}$) to represent the class of the image.

\textbf{Transformer encoder}. The encoder consists of a series of ViT blocks, where each block comprises two sub-modules: Multi-Head Attention (\texttt{MHA}) and Multi-Layer Perceptron (\texttt{MLP}). In each \texttt{MHA} module, we applied the standard self-attention formulation~\citep{vaswani2017attention}: $\text{Attention}(Q, K, V) = \text{softmax}(QK^T/\sqrt{d})V$, where query $Q$, key $K$ and value $V$ are linear projections of the input $z_b$ at block $b$. Conceptually, the self-attention layer assigns a pairwise attention value among all the patches (or tokens). In addition to the standard formulation, we also experimented with the Locality Self Attention (LSA, \citealt{lee2021vision}), where a diagonal mask is applied to $QK^T$ to prevent strong connections in self-tokens (i.e. diagonal values in $QK^T$), thus improving the locality inductive bias. Each sub-module is preceded by Layer Normalization (\texttt{LayerNorm}, \citealt{ba2016layer}), and followed by a residual connection to the next module.

\textbf{Reshape representation}. To make the dimensions compatible with the Gaussian readout module (see Section~\ref{previous-work} for an overview), we reshape the 2d core output $z \in \mathbb{R}^{l \times d}$ to $\mathbb{R}^{d \times h' \times w'}$, where $l = h' \times w'$ and $h' \leq w'$. Note that if the number of patches $l$ is not sufficiently large, it is possible for the same position in $z$ to be mapped to multiple neurons, which could lead to adverse effects. For instance, in the extreme case of $l = 1$, all neurons would be mapped to a single $p \times p$ region in the visual stimulus (i.e. they would have the same visual receptive field), which is not biologically plausible given the size of the recorded cortical area~\citep{garrett2014topography}. We therefore set the stride size of the patch encoder as a hyperparameter and allow for overlapping patches, thus letting the hyperparameter optimization algorithm select the optimal number of patches. Given $x_\text{image} \in \mathbb{R}^{c \times h=36 \times w=64}$, the V{\small 1}T core has an output dimension of $\mathbb{R}^{d \times h'=29 \times w'=57}$.

\subsubsection{Incorporating behaviors} \label{methods:v1t-core} 

Previous studies have shown that visual responses can be influenced by behavioral variables and brain states; for example, changes in arousal, which can be monitored by tracking pupil dilation, lead to stronger (or weaker) neural responses~\citep{reimer2016pupil, larsen2018neuromodulatory}. As a consequence, the visual representation learned by the core module should also be adjusted according to the brain state. Here, instead of inputting a vector that is a concatenation of the visual stimulus $x_\text{image} \in \mathbb{R}^{c \times h \times w}$ and behavioral information $x_\text{behaviors} \in \mathbb{R}^v$ in the channel dimension (i.e. $\mathbb{R}^{(c + v) \times h \times w}$, see Section~\ref{previous-work}), we propose an alternative method to integrate behavioral variables with the visual stimulus using a novel ViT-based architecture -- V{\footnotesize 1}T, illustrated in Figure~\ref{figure:v1t-architecture}.

We introduced a behavior MLP module ($\texttt{B-MLP}: \mathbb{R}^v \rightarrow \mathbb{R}^d$) at the beginning of the encoder block which learns to adjust the visual latent vector $z$ based on the observed behavioral states $x_\text{behaviors}$. Each \texttt{B-MLP} module comprises two fully-connected layers with $d$ hidden units and a dropout layer in between; $\tanh$ activation is used so that the adjustments to $z$ can be both positive and negative. Importantly, as layers in DNNs learn different features of the input, usually increasingly abstract and complex with deeper layers~\citep{zeiler2014visualizing, raghu2021vision}, we hypothesize that the influence of the internal brain state should therefore change from layer to layer. To that end, we learned a separate $\texttt{B-MLP}_b$ at each block $b$ in the V{\small 1}T core, thus allowing level-wise adjustments to the visual latent variable. Formally, $\texttt{B-MLP}_b$ projects $x_\text{behaviors}$ to the same dimension of the embeddings $z_{b - 1}$, followed by an element-wise summation between latent behavioral and visual representations, and then the rest of the operations in the encoder block:
\begin{align}
    z_b &\leftarrow z_{b - 1} + \texttt{B-MLP}_b(x_\text{behaviors}) \\
    z_b &\leftarrow \texttt{MHA}_b(\texttt{LayerNorm}(z_b)) + z_b \\
    z_b &\leftarrow \texttt{MLP}_b(\texttt{LayerNorm}(z_b)) + z_b
\end{align}
where $z_0$ denotes the original patch embeddings. To compare the prediction performance difference due to our proposed behavior module, we also trained an equivalent Vision Transformer (denoted as ViT) with the same architecture as V{\small 1}T except that it integrates behavioral information in the same manner as the CNN model (i.e. ViT inputs $\mathbb{R}^{(c + v) \times h \times w}$).

\subsection{Training and evaluation} \label{methods:training-and-evaluation}

In order to isolate the change in prediction performance that is solely due to the proposed core architectures, we employed the same readout architectures by \citet{lurz2021generalization}, as well as a similar data preprocessing and model training procedure. We used the same train, validation and test split provided by the two datasets (see Section~\ref{neural-data}). Natural images, recorded responses, and behavioral variables (i.e. pupil dilation, dilation derivative, pupil center, running speed) were standardized using the mean and standard deviation measured from the training set and the images were then resized to $36 \times 64$ pixels from $144 \times 256$ pixels. The shared core and per-animal readout modules were trained jointly using the AdamW optimizer~\citep{loshchilov2018decoupled} to minimize the Poisson loss 
\begin{align}
    \mathcal{L}^\text{Poisson}_m(r, o) = \sum^{n_t}_{t=1}\sum^{n_m}_{i=1}\Big(o_{i,t} - r_{i,t}\log(o_{i,t})\Big)
\end{align}
between the recorded responses $r$ and predicted responses $o$, where $n_t$ is the number of trials in one batch and $n_m$ the number of neurons for mouse $m$. A small value $\varepsilon = 1e-8$ was added to both $r$ and $o$ prior to the loss calculation to improve numeric stability. Gradients from each mouse were accumulated before a single gradient update to all modules. We tried to separate the gradient update for each animal, i.e. one gradient update per core-readout combination, but this led to a significant drop in performance. We suspect this is because the core module failed to learn a generalized representation among all animals when each update step only accounted for gradient signals from one animal. We used a learning rate scheduler in conjunction with early stopping: if the validation loss did not improve over 10 consecutive epochs, we reduced the learning rate by a factor of $0.3$; if the model still had not improved after 2 learning rate reductions, we then terminated the training process. Dropout~\citep{srivastava2014dropout}, stochastic depths~\citep{huang2016deep}, and L1 weight regularization were added to prevent overfitting. The weights in dense layers were initialized by sampling from a truncated normal distribution ($\mu = 0.0, \sigma = 0.02$), where the bias values were set to 0.0; whereas the weight and bias in \texttt{LayerNorm} were set to 1.0 and 0.0. Each model was trained on a single Nvidia RTX 2080Ti GPU and all models converged within 200 epochs. Finally, we employed Hyperband Bayesian optimization~\citep{li2017hyperband} to find the hyperparameters that achieved the best performance in the validation set. This included finding the optimal tokenization method and self-attention mechanism. The initial search space and final hyperparameter settings are detailed in Table~\ref{table:v1t_hyperparameter}. We independently performed a hyperparameter search on the CNN model, though we failed to find a configuration that achieves better performance than the settings provided by \citet{lurz2021generalization} and \citet{franke2022state}. While learning rate warm-up and pre-training on large datasets are considered the standard approach to train Transformers~\citep{xiong2020layer, han2022survey}, in order to stay consistent with previous work and to isolate the performance gain solely due to the architectural change, all models presented in this work are trained from scratch and follow the same procedure stated above.

The prediction performance of our models was measured by the single trial correlation metric, used by \citet{willeke2022sensorium} and \citet{franke2022state}, which can also account for the trial-to-trial variability in the test set where the same visual stimuli were shown multiple times. We computed the correlation between recorded $r$ and predicted $o$ responses:
\begin{align}
    \text{trial corr.}(r, o) = \frac{\sum_{i,j}(r_{i,j} - \bar{r})(o_{i,j} - \bar{o})}{\sqrt{\sum_{i,j}(r_{i,j} - \bar{r})^2\sum_{i,j}(o_{i,j} - \bar{o})^2}}
\end{align}
where $\bar{r}$ and $\bar{o}$ are the average recorded and predicted responses across all trials in the test set. 

\section{Results} \label{results}

Here, we first discuss the final core architecture chosen after the Bayesian hyperparameter optimization, followed by a comparison of our proposed core against baseline models on the two large-scale mouse V1 datasets. Moreover, we analyze the trained core module and present the insights that can be gained from it. We present the cross-animal and cross-dataset generalization in Appendix~\ref{appendix:generalization}.

\begin{table}[ht]
    \caption{The single trial correlation (\textsc{corr.}) between predicted and recorded responses in \textsc{Dataset S} and \textsc{Dataset F} test set. $\Delta\text{CNN}$ and $\Delta\text{V\textup{i}T}$ show the relative differences against the CNN~\citep{lurz2021generalization} and ViT models with behavior variables; we additionally fitted a CNN and ViT core with stimulus-response pairs (\textsc{behav}: \raisebox{-0.6pt}{\twemoji{multiply}}) to evaluate the prediction performance without behavioral information. \textsc{sd} shows the standard deviation across animals and detailed per-animal results are available in Appendix~\ref{appendix:prediction-results}.} \label{table:average-results}
    \begin{center}
    \begin{small}
    \begin{sc}
    \begin{tabular}{lccrr|crr}
        \toprule
         & & \multicolumn{3}{c}{\textsc{Dataset S} (\citeauthor{willeke2022sensorium})} & \multicolumn{3}{c}{\textsc{Dataset F} (\citeauthor{franke2022state})} \\
         & behav & corr. (sd) & \multicolumn{1}{c}{$\Delta\text{CNN}$} & \multicolumn{1}{c}{$\Delta\text{V\textup{i}T}$} & corr. (sd) & \multicolumn{1}{c}{$\Delta\text{CNN}$} & \multicolumn{1}{c}{$\Delta\text{V\textup{i}T}$} \\
        \midrule
        LN & \twemoji{check mark} & 0.275 (0.019) & -27.2\% & -33.7\% & 0.223 (0.040) & -28.0\% & -35.4\%\\
        CNN & \twemoji{multiply} & 0.300 (0.021) & -20.6\% & -27.6\% & & & \\
        CNN & \twemoji{check mark} & 0.378 (0.029) & 0.0\% & -8.7\% & 0.309 (0.070) & 0.0\% & -10.3\% \\
        V\textup{i}T & \twemoji{multiply} & 0.319 (0.024) & -15.6\% & -22.9\% & & & \\
        V\textup{i}T & \twemoji{check mark} & 0.414 (0.032) & +9.5\% & 0.0\% & 0.344 (0.041) & +11.4\% & 0.0\% \\
        V{\scriptsize 1}T & \twemoji{check mark} & \textbf{0.426} (0.027) & +12.7\% & +3.0\% & \textbf{0.368} (0.032) & +19.1\% & +6.9\% \\
        \midrule
        \multicolumn{8}{l}{Ensemble of 5 models} \\
        \midrule
        CNN & \twemoji{check mark} & 0.404 (0.025) & +6.9\% & -2.3\% & 0.340 (0.050) & +10.0\% & -1.3\% \\
        V\textup{i}T & \twemoji{check mark} & 0.424 (0.026) & +12.2\% & +2.4\% & 0.365 (0.037) & +18.1\% & +6.0\%\\
        V{\scriptsize 1}T & \twemoji{check mark} & \textbf{0.439} (0.027) & +16.1\% & +6.1\% & \textbf{0.378} (0.033) & +22.3\% & +3.8\% \\
        \bottomrule
    \end{tabular}
    \end{sc}
    \end{small}
    \end{center}
\end{table}

\textbf{V{\small 1}T benefits from smaller and overlapping patches}\label{results:v1t-benefits-from-smaller-and-overlapping-patches}. We first looked at how hyperparameters of ViT and V{\small 1}T affect model performance. We observed the predictive performance to be quite sensitive towards number of patches, patch size and patch stride. The most performant models used a patch size of 8 and a stride size of 1, thus extracting the maximum number of patches. We note that this allows the readout to learn a mapping from the shared core representation of the stimulus to the cortical position of each neuron that spans across the whole image, and not just a part of the image. Since the visual receptive fields of neurons are distributed across a large area of the monitor given the size of the recorded cortical area, this leads to more accurate response predictions from the model. Furthermore, we found that the two efficient tokenizers, SPT and CCT, whose aim is to reduce the number of patches, both failed to improve the model performance, reiterating that a finer tiling of the image is crucial for accurate predictions of cortical activity. Moreover, we found that the LSA attention mechanism, which encourages the model to learn from inter-tokens by masking out the diagonal self-token, led to worse performance, suggesting information from adjacent patches in this task is not as influential as it is in image classification. Appendix~\ref{appendix:hyperparameters} details the importance of each hyperparameter and the test performance trade-off among the various tokenizers and attention mechanisms. Lastly, we found that V{\small 1}T with layer-wise \texttt{B-MLP} modules yields the best results, indicating that the modulation introduced by behavioral information varies as the core learns different visual representations at deep layers. Further analysis and discussion on the \texttt{B-MLP} module are presented in Appendix~\ref{appendix:bmlp-activation}.

\textbf{V{\small 1}T outperforms CNN}\label{results:v1t-outperforms-cnn}. Next, we compared the tuned ViT and V{\small 1}T cores against a baseline linear-nonlinear (LN) model and the previous state-of-the-art CNN model~\citep{lurz2021generalization} on the two large scale mouse V1 datasets (see Section~\ref{neural-data}). We also trained a CNN and ViT core on response-stimuli pairs only on \textsc{Dataset S}, to evaluate the importance of behavioral information in response predictions. Table~\ref{table:average-results} summarizes the test performance on the two datasets, results of per-animal performance and an alternative metric are available in Appendix~\ref{appendix:prediction-results}. By simply replacing the CNN core module with the tuned ViT architecture, we observed a considerable improvement in response predictions across all animals, with an average increase of $9.5\%$ and $11.3\%$ in single trial correlation over the CNN model in \textsc{Dataset S} and \textsc{Dataset F} respectively. Thus far, the core module encoded the brain state of the animals by concatenating behavioral variables as additional channels in the natural image. With that said, our proposed V{\small 1}T core, which encodes the brain state via the \texttt{B-MLP} nonlinear transformations, further improved the average prediction performance by $2.9\%$ and $7.0\%$ in the two datasets, or $12.7\%$ and $19.1\%$ over the CNN model.

As demonstrated in the Sensorium Challenge~\citep{sensorium2022presentation} and \citet{franke2022state}, ensemble learning is a common approach to improve neural predictive models. Following the procedure in \citet{franke2022state}, we trained 10 models with different random seeds and selected the 5 best models based on their validation performance. The average of the selected models constituted the output of the ensemble model. The CNN ensemble model achieved an average improvement of $6.9\%$ in \textsc{Dataset S} as compared to its non-ensemble variant. Nevertheless, the individual V{\small 1}T model still outperformed the CNN ensemble by $5.4\%$. A V{\small 1}T ensemble trained with the same procedure achieved an average single trial correlation of 0.439, which corresponds to an $8.7\%$ improvement over the CNN ensemble model. Altogether, our proposed core architecture set a new benchmark in both gray-scale and colored visual response prediction.

\begin{figure}[ht]
    \centering
    \includegraphics[width=0.85\linewidth]{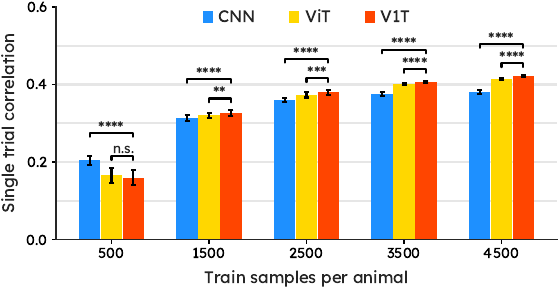}
    \caption{Prediction performance when trained with 500, 1500, 2500, 3500 and (all) 4500 samples per animal in \textsc{Dataset S}. The models were each trained with 30 different random seeds. The error bar shows the standard deviations of the repeated experiments, and the statistical difference (two-sided t-test) in CNN vs V{\small 1}T and ViT vs V{\small 1}T in each sample group is shown above each pair of bars (****:~$p \leq 0.0001$).}
    \label{figure:limit_data}
\end{figure}

\textbf{Sample efficiency}. Most neural datasets are constrained by their limited size, due to technical and/or ethical limitations, while typical DNNs require a large amount of data to train on, especially Transformer-based models~\citep{han2022survey}. Here, we evaluate the sample efficiency of the CNN, ViT and V{\small 1}T models by fitting them with 500 (11\%), 1500 (33\%), 2500 (55\%), 3500 (77\%) and 4500 (100\%) samples per animal in \textsc{Dataset S}~\citep{willeke2022sensorium}. Figure~\ref{figure:limit_data} shows the single trial correlation in the test set for the three models trained on different sample sizes, each with 30 different random seeds. Overall, we found that V{\small 1}T outperforms the CNN model even at 1500 training samples per animal. Moreover, the predictive performance of the CNN model plateaus at around 3500 training samples, while V{\small 1}T keeps improving, suggesting that the ViT-based model can continue to improve with more data.

\begin{figure}[ht]
    \centering{
        \phantomsubcaption\label{figure:aRF_sample}
        \phantomsubcaption\label{figure:aRF_sigma_distributions}
        \phantomsubcaption\label{figure:validation_attention}
        \phantomsubcaption\label{figure:test_attention}
    }
    \centering
    \includegraphics[width=1\linewidth]{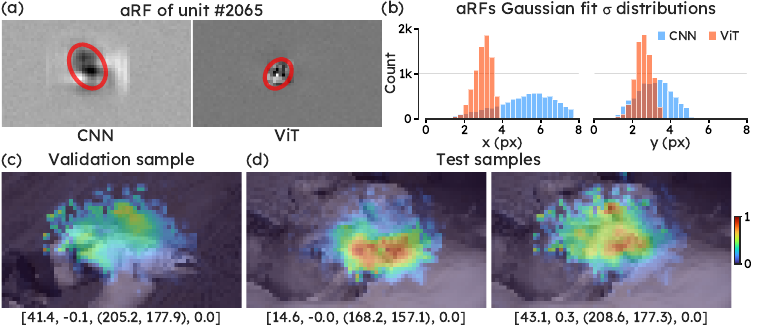}
    \caption{(\subref{figure:aRF_sample}) Estimated artificial receptive field (aRF) and 2d Gaussian fit (red circle shows 1 standard deviation ellipse) of the same artificial unit from the CNN and ViT models trained without behaviors. Visually, the ViT learns narrower aRFs, more examples in Appendix~\ref{appendix:artificial-receptive-fields}. To quantify the size of the aRFs, we compared the fitted Gaussian over all units from \textsc{Mouse A}; (\subref{figure:aRF_sigma_distributions}) the distributions of the standard deviations shows that the ViT learns notably narrower aRFs. V{\small 1}T attention visualization on \textsc{Mouse A} (\subref{figure:validation_attention}) validation and (\subref{figure:test_attention}) test samples. Each attention map was normalized to $[0, 1]$, and the behavioral variables of the corresponding trial are shown below the image in the format of [pupil dilation, dilation derivative, pupil center ($x$, $y$), speed]. More examples in Appendix~\ref{appendix:attention-rollout-maps}. }
    \label{figure:attention_and_aRFs}
\end{figure}

\textbf{Spatial tuning difference}\label{results:spatial-tuning-difference}. As expected, models trained without behavioral information led to worse results (see \textsc{behavior}: \raisebox{-0.6pt}{\twemoji{multiply}} in Table~\ref{table:average-results}). Nevertheless, we observed an average $6.3\%$ improvement in stimuli-response prediction with the tuned ViT core over the CNN model in \textsc{Dataset S}. To further our understanding of why the ViT might be performing better in visual response prediction, we evaluated the discrepancies in spatial tuning of the two models by comparing their artificial receptive fields (aRFs). Appendix~\ref{appendix:artificial-receptive-fields} details the procedure. Briefly, we presented the models with thousands of white noise images and then summed the images weighted by the response prediction to estimate the aRF of each artificial unit. Figure~\ref{figure:aRF_sample} shows the aRF of the same artificial unit from the CNN and ViT model. Visually, the aRFs of the ViT model appear to be narrower and qualitatively different from the aRFs of the CNN. In order to quantify the aRF sizes, we fitted a 2d Gaussian to each aRF and observed a significant difference in the standard deviation distributions, shown in Figure~\ref{figure:aRF_sigma_distributions}. Overall, the aRFs of the ViT model have a much narrower spread, with a mean standard deviation of $3.0 \pm 0.5$ and $2.6 \pm 0.4$ in the horizontal and vertical directions over all artificial units, considerably lower than the $5.1 \pm 1.5$ and $3.1 \pm 0.9$ of the CNN. These results show that the artificial units in the CNN and ViT learn notably different aRFs. Given that we did not constrain the aRF size, our results suggest that the narrower fields allow ViT to learn location-dependent features that are beneficial for visual response prediction.

\textbf{Self-attention visualization}\label{results:self-attention-visualization}. In addition to the performance gain in the proposed core architecture, the self-attention mechanism inherent in Transformers can be used to visualize areas in the input image that the model learns to focus on. In our case, it allows us to detect the regions in the visual stimulus that drive the neural responses. To that end, we extracted the per-stimulus attention map learned by the V{\small 1}T core module via Attention Rollout~\citep{abnar2020quantifying, dosovitskiy2021an}. Briefly, we aggregated the attention weights (i.e. $\text{Softmax}(QK^T/\sqrt{d})$) across all heads in \texttt{MHA}, and then multiplied the weights over all layers (blocks), recursively. Figure~\ref{figure:attention_and_aRFs} shows the normalized average attention weights superimposed to the input images from \textsc{Mouse A}, with more examples available in Appendix~\ref{appendix:attention-rollout-maps}. Given that the position of the computer monitor was chosen in order to center the population receptive field, V1 responses from the recorded region should be mostly influenced by the center of the image~\citep{willeke2022sensorium}. Here, we can see a clear trend where the core module is focusing on the central regions of the images to predict the neural response, which aligns with our expectation from the experiment conditions. Interestingly, when the core module inputs the same image but with varying behaviors (i.e. Figure~\ref{figure:test_attention}), we noticed variations in the attention patterns. This suggests that the V{\small 1}T core is able to take behavioral variables into consideration and adjust its attention solely based on the brain state.

These attention maps can inform us of the area of the image (ir)relevant for triggering visual neuronal responses which, in turn, allow us to build more sophisticated predictive models. For instance, the core module consistently assigned higher weights to patches in the center of the image, suggesting information at the edges of the image are less (or not at all) relevant for the recorded group of neurons. As a practical example, we eliminated irrelevant information in the stimuli by center cropping the image to $\alpha 144 \times \alpha 256$ pixels where $0 < \alpha \leq 1$, prior to downsampling the input to $36 \times 64$ pixels. We found that a crop factor of $\alpha = 0.8$ (i.e. removing $36\%$ of the total number of pixels) further improved the single trial correlation to $0.430$, or $13.8\%$ better than the CNN. Note that we also obtained similar improvement with the CNN model. 

\textbf{Self-attention correlates with pupil center}\label{results:self-attention-correlates-with-pupil-center}. To further explore the relationship between the attention weights learned by the core module and the behavioral information, we measured the absolute correlation between the center of mass of the attention maps and the pupil centers in the vertical and horizontal axes. The correlation coefficient of each animal in \textsc{Dataset S} is summarized in Table~\ref{table:pupil-correlation}. Overall, we found a moderate correlation between the attention maps and the pupil center of the animal, with an average correlation (standard deviation) of $0.525$ ($0.079$) and $0.409$ ($0.105$) in the horizontal and vertical directions across animals. This relationship demonstrates that the attention maps can reveal the impact of behavioral variables on the neural responses. Therefore, this framework can be particularly useful for studies investigating the coding of visual information across visual cortical areas (V1 and higher visual areas), as the model could determine what part(s) of the visual stimulus is processed along the ``hierarchy'' of visual cortical areas. Since higher visual areas are known to have larger receptive fields~\citep{wang2007area, glickfeld2014mouse}, we would expect a larger part of the image to be relevant for the core module. Further investigation of the attention map could also be used to determine which part of a visual scene was relevant when performing more specific tasks, such as object recognition, decision-making, or spatial navigation. 

\begin{table}[ht]
    \caption{Correlations between the center of mass of the attention maps and pupil centers in the (x-axis) horizontal and (y-axis) vertical direction in \textsc{Dataset S} test set, all with a p-value $\ll 0.0001$.} \label{table:pupil-correlation}
    \begin{center}
    \begin{small}
    \begin{sc}
    \begin{tabular}{>{\centering\arraybackslash}p{0.1\textwidth}>{\centering\arraybackslash}p{0.2\textwidth}>{\centering\arraybackslash}p{0.2\textwidth}}
        \toprule
        Mouse & x-axis & y-axis \\
        \midrule
        A & 0.682 (****) & 0.568 (****) \\
        B & 0.489 (****) & 0.493 (****) \\
        C & 0.505 (****) & 0.370 (****) \\
        D & 0.484 (****) & 0.310 (****) \\
        E & 0.464 (****) & 0.302 (****) \\
        \bottomrule
    \end{tabular}
    \end{sc}
    \end{small}
    \end{center}
\end{table}

\section{Discussion} \label{discussion}

In this work, we presented a novel core architecture V{\small 1}T to model the visual and behavioral representations of mouse V1 activities in response to natural visual stimuli. The model outperformed the previous state-of-the-art CNN~\citep{lurz2021generalization} model on two large-scale mouse V1 datasets by a considerable margin (12.7\% and 19.1\%). In contrast to the winning submissions at the Sensorium Challenge~\citep{sensorium2022presentation}, which focused on data augmentation and building large ensembles based on the CNN model, we instead introduced a new architecture as the shared core module. Our best model achieved a single trial correlation of $0.428$ and $0.444$ (correlation to average: $0.634$ and $0.650$) in the two held-out test sets, which would place us 2\textsuperscript{nd} place in the leaderboard, and the best method across all models not taking the neuronal response trends over time into account. In addition, we also showed that V{\small 1}T can be competitive in the low data regime, and that its performance continues to improve with more data to a larger extend than the CNN model. To the best of our knowledge, our approach is also the first ViT-based model to outperform CNNs in mouse V1 response prediction.

With a strong neural predictive performance, this model also provides a framework to investigate \textit{in silico} the computations in the visual system, and in particular, the modulation of neural responses by behavioral variables.
In this study, we included speed of the animal in the virtual corridor, pupil dilation, dilation derivative and pupil center as behavioral variables. For each of these variables, there is prior evidence showing that they do affect responses in V1. For instance, \citet{pakan2018action} showed that 12\% of the recorded V1 neurons decreased their activity with lower running speed, suggesting a clear benefit of considering the speed of the animal for predicting V1 responses. Pupil dilation has been shown to be related to arousal of the animal, with complex modality dependent effects of arousal on the mouse sensory cortex~\citep{shimaoka2018effects}. The pupil center represents the fixation point of the animal and is a proxy for what the animal is paying attention to. As a proof of principle of how a Vision Transformer can be used to gain insights into the importance of behavioral variables for V1 responses, we showed that the center of the self-attention maps learned by our model correlates with the pupil center of the animals, highlighting how features of this architecture do reflect properties of cortical neurons' receptive fields, in this case, the retinotopy. Moreover, our model is able to exploit certain anatomical information, for example the location of neurons within the primary visual cortex, from which we can roughly infer the location of their receptive field since the retinotopic map of mouse primary visual cortex is well characterized~\citep{zhuang2017extended}. However, while the CNN architecture was inspired by receptive fields of the visual cortex~\citep{fukushima1980neocognitron}, the Vision Transformer architecture was not and has no direct biological counterpart. Therefore, it is challenging to map the abstract components of a Vision Transformer onto the anatomy or biophysics of the brain.

Nevertheless, the V{\small 1}T model has a number of limitations. Firstly, only one-dimensional behavioral information can be incorporated since the model integrates scalars into the latent embedding via the \texttt{B-MLP} module. Additional architecture engineering is needed if the behavioral variables have varying (and higher) dimensions, for instance, 3D poses~\citep{mathis2018deeplabcut}. Secondly, in the case of very limited data (e.g. 500 samples, see Figure~\ref{figure:limit_data}), CNN-based models are likely to outperform ViTs, which typically require considerable amount of data to be performant~\citep{han2022survey}.

In future work, we plan to further investigate the relationship between behavioral variables and neural responses. The attention visualization technique, for instance, enables ablation studies on the effect of each behavioral variable, such as pupil dilation or running speed, on the neural activity. Moreover, we plan to extend the method to recordings of the visual cortex in response to natural videos, to track how this relationship may evolve over time, as well as experiments in naturalistic settings, to know which part of a visual scene is relevant for certain behaviors.


\subsubsection*{Acknowledgments}
We sincerely thank \citeauthor{willeke2022sensorium} and \citeauthor{franke2022state} for making their high-quality large-scale mouse recordings publicly available which make this work possible. We would also like to thank Antonio Vergari, Matthias Hennig and Robyn Greene for their insightful comments and suggestions on improving the manuscript. BML was supported by the United Kingdom Research and Innovation (grant EP/S02431X/1), UKRI Centre for Doctoral Training in Biomedical AI at the University of Edinburgh, School of Informatics. This project has received funding from the European Research Council (ERC) under the European Union’s Horizon 2020 research and innovation programme (grant agreement No. 866386; to N.R.). For the purpose of open access, the author has applied a creative commons attribution (CC BY) licence to any author accepted manuscript version arising.

\bibliography{reference}
\bibliographystyle{apalike}

\clearpage

\counterwithin{figure}{section}
\counterwithin{table}{section}

\appendix

\section{Appendix}

\begin{table}[ht]
    \caption{Experimental information of \textsc{Mouse A} to \textsc{E} from \textsc{Dataset S}~\citep{willeke2022sensorium} and \textsc{Mouse F} to \textsc{O} from \textsc{Dataset F}~\citep{franke2022state}. Each mouse has a unique recording ID (column 2) although we assigned a separate mouse ID (column 1) to use throughout this paper for simplicity.} \label{table:recording-info}
    \begin{center}
    \begin{small}
    \begin{sc}
    \begin{tabular}{clcccc}
        \toprule
        Mouse & rec. ID & num. neurons & total trials & num. test \\
        \midrule
        A & 21067-10-18 & 8372 & 5994 & 998 \\
        B & 22846-10-16 & 7344 & 5997 & 999 \\
        C & 23343-5-17 & 7334 & 5951 & 989 \\
        D & 23656-14-22 & 8107 & 5966 & 993 \\
        E & 23964-4-22 & 8098 & 5983 & 994 \\
        \midrule
        F & 25311-10-26 & 867 & 7358 & 1475 \\
        G & 25340-3-19 & 922 & 7478 & 1497 \\
        H & 25704-2-12 & 773 & 7500 & 1500 \\
        I & 25830-10-4 & 1024 & 7360 & 1473 \\
        J & 26085-6-3 & 910 & 7464 & 1495 \\
        K & 26142-2-11 & 1121 & 7500 & 1500 \\
        L & 26426-18-13 & 1125 & 7500 & 1500 \\
        M & 26470-4-5 & 1160 & 7473 & 1495 \\
        N & 26644-6-2 & 824 & 7500 & 1500 \\
        O & 26872-21-6 & 1109 & 7466 & 1495 \\
        \bottomrule
    \end{tabular}
    \end{sc}
    \end{small}
    \end{center}
\end{table}

\clearpage

\subsection{Hyperparameters}~\label{appendix:hyperparameters}

\begin{table}[ht]
    \caption{ViT and V{\footnotesize 1}T cores - Gaussian readout hyperparameter search space and their final settings after a Hyperband Bayesian optimization~\citep{li2017hyperband}.} \label{table:v1t_hyperparameter}
    \begin{center}
    \begin{small}
    \begin{sc}
    \begin{tabular}{lcr}
        \toprule
        hyperparameter & search space & final value \\
        \midrule
        Core & & \\
        num. blocks & uniform, min: 1, max: 8 & 4 \\
        num. heads & uniform, min: 1, max: 12 & 4 \\
        patch size & uniform, min: 2, max: 16 & 8 \\
        patch stride & uniform, min: 1, max: patch size & 1 \\
        patch method & sliding window, 2d conv, SPT, CCT & sliding window \\
        patch dropout & uniform, min: 0, max: 0.5 & 0.0229 \\
        embedding size & uniform, min: 8, max: 1024, interval: 1 & 155 \\
        mha method & original, LSA & original \\
        mha dropout & uniform, min: 0, max: 0.5 & 0.2544 \\
        mlp size & uniform, min: 8, max: 1024, interval: 1 & 488 \\
        mlp dropout & uniform, min: 0, max: 0.5  & 0.2544 \\
        stochastic depth dropout & uniform, min: 0, max: 0.5 & 0.0 \\
        L1 weight regularization & uniform, min: 0, max: 1 & 0.5379 \\
        initial learning rate & uniform, min: 0.005, max: 0.0001 & 0.0016 \\
        \midrule
        Readout & & \\
        position network num. layers & uniform, min: 1, max: 4, interval: 1 & 1 \\
        position network num. units & uniform, min: 2, max: 128, interval: 2 & 30 \\
        bias initialization & 0, mean standardized response & 0 \\
        L1 weight regularization & uniform, min: 0, max: 1 & 0.0076 \\
        \bottomrule
    \end{tabular}
    \end{sc}
    \end{small}
    \end{center}
\end{table}

\begin{table}[ht]
    \caption{ViT and V{\footnotesize 1}T hyperparameter importance in Hyberhand Bayesian Optimization~\citep{li2017hyperband} via Weights \& Biases~\citep{wandb}. \textsc{importance} shows the degree to which the hyperparameter is useful to predict the evaluation metric (e.g. single trial correlation in the validation set) and \textsc{correlation} shows the linear correlation between the hyperparameter and the evaluation metric. Details on the calculation and interpretation of the hyperparameter importance and correlation are available at \href{https://docs.wandb.ai/guides/app/features/panels/parameter-importance}{docs.wandb.ai/guides/app/features/panels/parameter-importance}.} \label{table:hyperparameter_importance}
    \begin{center}
    \begin{small}
    \begin{sc}
    \begin{tabular}{lcc}
        \toprule
        hyperparameter & importance & correlation \\
        \midrule
        embedding size & 0.393 & -0.626 \\
        patch stride & 0.164 & -0.358 \\
        patch size & 0.111 & -0.297 \\
        initial learning rate & 0.046 & 0.279 \\
        L1 weight regularization & 0.030 & -0.242 \\
        num. blocks & 0.030 & 0.093 \\
        num. heads & 0.028 & -0.070 \\
        batch size & 0.026 & -0.093 \\
        mha dropout & 0.025 & -0.034 \\
        patch method & 0.024 & -0.174 \\
        mlp dropout & 0.022 & 0.133 \\
        mlp size & 0.019 & -0.186 \\
        stochastic depth dropout & 0.019 & -0.225 \\
        patch dropout & 0.017 & -0.105 \\
        mha method & 0.014 & 0.001 \\
        \bottomrule
    \end{tabular}
    \end{sc}
    \end{small}
    \end{center}
\end{table}

\begin{table}[ht]
    \caption{Best prediction performance in single trial correlation (standard deviation across animals) on Dataset S with respect to choice of attention formulation and patch/tokenization method. \textsc{original} denotes the original self-attention formulation by \citealt{vaswani2017attention} and \textsc{LSA} denotes the Locality Self Attention mechanism proposed by \citealt{lee2021vision}. \textsc{SPT} denotes Shifted Patch Tokenization~\citep{lee2021vision} and \textsc{CCT} denotes the tokenization method introduced in Compact Convolution Transformer~\citep{hassani2021escaping}. Section~\ref{methods:model} details the model architectural differences and Section~\ref{results:v1t-benefits-from-smaller-and-overlapping-patches} discusses their prediction results.} \label{table:mha_patch_methods}
    \begin{center}
    \begin{small}
    \begin{sc}
    \begin{tabular}{c|cccc}
        \toprule
        \diagbox[dir=NW]{MHA}{patch}\quad method & sliding window & 2d conv & spt & cct \\
        \midrule
        original & \textbf{0.426} (0.027) & 0.411 (0.022) & 0.406 (0.024) & 0.392 (0.026) \\
        lsa & 0.413 (0.023) & 0.415 (0.024) & 0.405 (0.024) & 0.385 (0.025) \\
        \bottomrule
    \end{tabular}
    \end{sc}
    \end{small}
    \end{center}
\end{table}

\subsection{\texttt{B-MLP} activation} \label{appendix:bmlp-activation}

\begin{figure}[ht]
    \centering
    \includegraphics[width=0.45\linewidth]{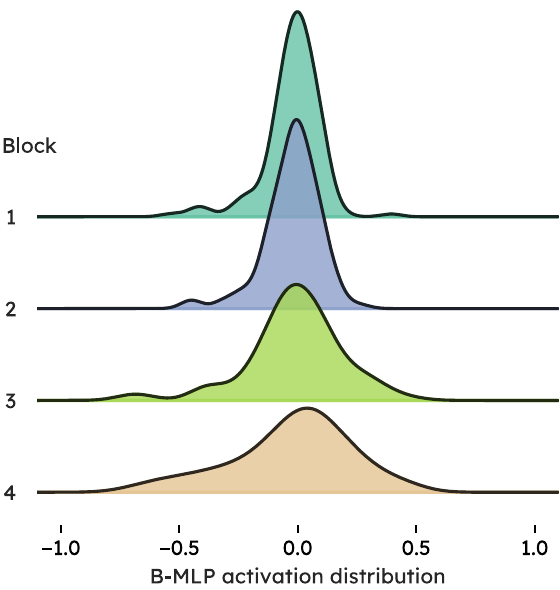}
    \caption{$\tanh$ activation distributions of \texttt{B-MLP} at each level (block) in the V{\footnotesize 1}T core. The spread of activation distributions indicates varying influence of behavioral variables at the block in the core module.}
    \label{figure:bmlp-activations}
\end{figure}

We investigated different variations of the \texttt{B-MLP} module. The motivation of the proposed behavior module is to enable the core to learn a shared representation of the visual and behavioral variables across the animals. Moreover, the level-wise connections allow the self-attention module in each V{\small 1}T block to encode different behavioral features with the latent visual representation. We experimented with a per-animal \texttt{B-MLP} module (while the rest of the core was still shared across animals) which did not perform any better than the shared counterpart, suggesting that the behavior module can indeed learn a shared internal brain state presentation. We also tested having the module in the first block only, as well as using the same module across all blocks (i.e. all $\texttt{B-MLP}_b$ shared the same weights). Both cases, however, led to worse results with a $2-4\%$ reduction in predictive performance on average. To further examine the proposed formulation, we analyzed the activation patterns of the shared behavior module at each level in V{\small 1}T, shown in Figure~\ref{figure:bmlp-activations}. We observed a noticeable distinction in \texttt{B-MLP} outputs in earlier versus deeper layers, with a higher spread in deeper layers, which corroborates our hypothesis that the influence of the behavioral variables differs at each level of the visual representation process.

\clearpage

\subsection{Prediction results} \label{appendix:prediction-results}

\begin{table}[ht]
    \caption{Single trial correlation between predicted and recorded responses in \textsc{Dataset S} test set. All models were trained with behaviors. To demonstrate that the extracted attention maps can inform us about the (ir)relevant regions in the visual stimulus, we trained an additional V{\footnotesize 1}T core with images center cropped to $\alpha h \times \alpha w$ pixels (See Section~\ref{results:self-attention-visualization}).} \label{table:sensorium-results}
    \vspace{-3mm}
    \begin{center}
    \begin{small}
    \begin{sc}
    \begin{tabular}{llllll|l}
        \toprule
         & \multicolumn{5}{c}{Mouse} & \\
         & \multicolumn{1}{c}{A} & \multicolumn{1}{c}{B} & \multicolumn{1}{c}{C} & \multicolumn{1}{c}{D} & \multicolumn{1}{c}{E} & \multicolumn{1}{c}{avg (sd)} \\
        \midrule
        LN & 0.262 & 0.306 & 0.281 & 0.263 & 0.262 & 0.275 (0.019) \\
        CNN & 0.350 & 0.424 & 0.385 & 0.371 & 0.360 & 0.378 (0.029) \\
        V\textup{i}T & 0.375 & 0.455 & 0.415 & 0.433 & 0.392 & 0.414 (0.032) \\
        V{\scriptsize 1}T & 0.401 & 0.464 & 0.430 & 0.436 & 0.401 & 0.426 (0.027) \\
        V{\scriptsize 1}T (center crop $\alpha = 0.8$) & \textbf{0.403} & \textbf{0.468} & \textbf{0.433} & \textbf{0.442} & \textbf{0.403} & \textbf{0.430} (0.028) \\
        \midrule
        \multicolumn{7}{l}{Ensemble of 5 models} \\
        \midrule
        CNN & 0.379 & 0.443 & 0.409 & 0.406 & 0.385 & 0.404 (0.025) \\
        V\textup{i}T & 0.398 & 0.460 & 0.421 & 0.440 & 0.401 & 0.424 (0.026) \\
        V{\scriptsize 1}T & \textbf{0.414} & \textbf{0.475} & \textbf{0.443} & \textbf{0.452} & \textbf{0.413} & \textbf{0.439} (0.027) \\
        \bottomrule
    \end{tabular}
    \end{sc}
    \end{small}
    \end{center}
\end{table}

\begin{table}[ht]
    \caption{Single trial correlation between predicted and recorded responses in \textsc{Dataset F} test set. All models were trained with behaviors.} \label{table:franke2022-results}
    \vspace{-3mm}
    \begin{center}
    \begin{small}
    \begin{sc}
    \resizebox{1\textwidth}{!}{
    \begin{tabular}{lllllllllll|l}
        \toprule
         & \multicolumn{10}{c}{Mouse} \\
         & \multicolumn{1}{c}{F} & \multicolumn{1}{c}{G} & \multicolumn{1}{c}{H} & \multicolumn{1}{c}{I} & \multicolumn{1}{c}{J} & \multicolumn{1}{c}{K} & \multicolumn{1}{c}{L} & \multicolumn{1}{c}{M} & \multicolumn{1}{c}{N} & \multicolumn{1}{c}{O} & \multicolumn{1}{c}{avg (sd)}\\
        \midrule
        LN & 0.194 & 0.254 & 0.214 & 0.279 & 0.255 & 0.233 & 0.148 & 0.231 & 0.174 & 0.243 & 0.223 (0.040)\\
        CNN & 0.253 & 0.371 & 0.184 & 0.377 & 0.329 & 0.319 & 0.207 & 0.331 & 0.341 & 0.376 & 0.309 (0.070)\\
        V\textup{i}T & 0.310 & 0.375 & 0.352 & 0.379 & 0.385 & 0.262 & 0.294 & 0.360 & 0.358 & 0.368 & 0.344 (0.041) \\
        V{\scriptsize 1}T & \textbf{0.326} & \textbf{0.386} & \textbf{0.387} & \textbf{0.394} & \textbf{0.398} & \textbf{0.373} & \textbf{0.298} & \textbf{0.377} & \textbf{0.363} & \textbf{0.379} & \textbf{0.368} (0.032)\\
        \midrule
        \multicolumn{12}{l}{Ensemble of 5 models} \\
        \midrule
        CNN & 0.268 & 0.383 & 0.341 & 0.393 & 0.347 & 0.336 & 0.242 & 0.345 & 0.355 & 0.388 & 0.340 (0.050)\\
        V\textup{i}T & 0.321 & 0.384 & 0.363 & 0.404 & 0.406 & 0.374 & 0.302 & 0.385 & 0.323 & 0.387 & 0.365 (0.037) \\
        V{\scriptsize 1}T & \textbf{0.336} & \textbf{0.397} & \textbf{0.391} & \textbf{0.406} & \textbf{0.408} & \textbf{0.383} & \textbf{0.306} & \textbf{0.388} & \textbf{0.373} & \textbf{0.392} & \textbf{0.378} (0.033) \\
        \bottomrule
    \end{tabular}
    }
    \end{sc}
    \end{small}
    \end{center}
\end{table}

\clearpage

\subsubsection{Correlation to Average}~\label{appendix:correlation_to_average}

Correlation to Average (\textsc{avg. corr.}) is another commonly used metric to evaluate neural predictive models~\citep{willeke2022sensorium}. It is the correlation between $r_{i, j}$ recorded and $o_{i,j}$ predicted responses over repeated $j$ trials of stimulus $i$ : 
\begin{align}
    \text{avg. corr.}(r, o) = \frac{\sum_{i}(\bar{r}_{i} - \bar{r})(o_{i} - \bar{o})}{\sqrt{\sum_{i}(\bar{r}_{i} - \bar{r})^2\sum_{i}(o_{i} - \bar{o})^2}}
\end{align}
where $\bar{r}_i = \frac{1}{J} \sum^{J}_{j=1} r_{i,j}$ is the average response across $J$ repeats, and $\bar{r}$ and $\bar{o}$ are the average recorded and predicted responses across all trials.

\begin{table}[ht]
    \caption{The Correlation to Average (\textsc{avg. corr.}) between predicted and recorded responses across all animals (\textsc{SD} shows the standard deviation) in \textsc{Dataset S} and \textsc{Dataset F} test sets. Table~\ref{table:average-results} shows the results in single trial correlation.} \label{table:correlation-to-average-results}
    \begin{center}
    \begin{small}
    \begin{sc}
    \begin{tabular}{lccrr|crr}
        \toprule
         & & \multicolumn{3}{c}{\textsc{Dataset S} (\citeauthor{willeke2022sensorium})} & \multicolumn{3}{c}{\textsc{Dataset F} (\citeauthor{franke2022state})} \\
         & behav & avg. corr. (SD) & \multicolumn{1}{c}{$\Delta\text{CNN}$} & \multicolumn{1}{c}{$\Delta\text{V\textup{i}T}$} & avg. corr. (SD) & \multicolumn{1}{c}{$\Delta\text{CNN}$} & \multicolumn{1}{c}{$\Delta\text{V\textup{i}T}$} \\
        \midrule
        LN & \twemoji{check mark} & 0.387 (0.023) & -33.1\% & -37.7\% & 0.312 (0.076) & -39.7\% & -42.5\% \\
        CNN & \twemoji{multiply} & 0.551 (0.024) & -4.7\% & -4.6\% & & &\\
        CNN & \twemoji{check mark} & 0.578 (0.027) & 0.0\% & -6.9\% & 0.516 (0.142) & 0.0\% & -4.7\% \\
        V\textup{i}T & \twemoji{multiply} & 0.568 (0.026) & -1.7\% & -8.5\% & & & \\
        V\textup{i}T & \twemoji{check mark} & 0.621 (0.030) & +7.4\% & 0.0\% & 0.542 (0.054) & 4.9\% & 0.0\% \\
        V{\scriptsize 1}T & \twemoji{check mark} & \textbf{0.629} (0.029) & +8.9\% & 1.4\% & \textbf{0.551} (0.022) & +6.6\% & +1.6\% \\
        \midrule
        \multicolumn{8}{l}{Ensemble of 5 models} \\
        \midrule
        CNN & \twemoji{check mark} & 0.610 (0.027) & +5.5\% & -1.7\%& \textbf{0.567} (0.050) & +9.9\% & +4.8\% \\
        V\textup{i}T & \twemoji{check mark} & 0.634 (0.027) & +9.7\% & +2.1\% & 0.566 (0.035) & +9.5\% & +4.4\% \\
        V{\scriptsize 1}T & \twemoji{check mark} & \textbf{0.644} (0.026) & +11.3\% & +3.7\% & 0.562 (0.023) & +8.9\% & +3.8\% \\
        \bottomrule
    \end{tabular}
    \end{sc}
    \end{small}
    \end{center}
\end{table}

\clearpage

\subsection{Cross-animal and cross-dataset generalization} \label{appendix:generalization}

DNN-based neural predictive models are often neuron/animal specific and do not generalize well to unseen neurons/animals.  Here, we evaluate generalization performance of CNN and V1T.

We first tested the cross-animal performance of the CNN and V{\small 1}T models by performing cross-validation over animals in \textsc{Dataset S}~\citep{willeke2022sensorium}. Specifically, we compare the model fitted on one animal (direct setting) against a model that was pre-trained on $N-1$ animals and whose readout was fine-tuned (with core frozen) on the left-out animal (transfer setting). We repeated this process for all 5 animals, and their results are summarized in Table~\ref{table:cross-animal-results}. On average, the V{\small 1}T model outperformed the CNN model by $3.3\%$ and $6.7\%$ in the direct and transfer settings, respectively. Moreover, the V{\small 1}T model experienced a larger level of performance gain in the transfer learning setting, with an average prediction improvement of $5.6\%$ over direct training, whereas the CNN had a $2.2\%$ gain. These results suggest that the V{\small 1}T core can generalize well to unseen animals, and also benefit from transfer learning to a greater extent.

Next, we evaluated the cross-dataset generalization performance. To that end, we fitted the models on the gray-scaled version (average channel dimension) of \textsc{Dataset F}~\citep{franke2022state}. We then froze the core module and trained the readouts on \textsc{Dataset S} and compared the loss in performance in this transfer setting. The results are presented in Table~\ref{table:cross-dataset-results} for the two core architectures. We observed a larger performance drop with the frozen V{\small 1}T model compared to the model trained directly, with an average deficit of $-19.0\%$, versus the $-12.9\%$ drop in the frozen CNN model. Similar to the cross-animal generalization, the CNN model exhibits a higher level of variation in prediction performance over the 5 animals. While the relative performance drop was greater for the V{\small 1}T core than for the CNN core, V{\small 1}T achieved better transfer results with an average single trial correlation of $0.345$, or about $4.9\%$ better than the frozen CNN ($0.329$).

\begin{table}[ht]
    \caption{\textbf{CNN vs V{\small 1}T cross-animal generalization in \textsc{Dataset S}}. We compare the test performance between (\textsc{direct}) fitting one model per animal and (\textsc{transfer}) pre-training a model on $N-1$ animals and fine-tuning the readout for the $N^\text{th}$ animal. We repeat the same leave-one-out process for all animals. $\Delta\textsc{direct}$ shows the relatively prediction performance of the \textsc{transfer} models over the \textsc{direct} models.} \label{table:cross-animal-results}
    \vspace{-3mm}
    \begin{center}
    \begin{small}
    \begin{sc}
    \begin{tabular}{llllll|lr}
        \toprule
         & \multicolumn{5}{c}{Mouse} & \\
         & \multicolumn{1}{c}{A} & \multicolumn{1}{c}{B} & \multicolumn{1}{c}{C} & \multicolumn{1}{c}{D} & \multicolumn{1}{c}{E} & \multicolumn{1}{c}{avg (sd)} & \multicolumn{1}{c}{$\Delta\textsc{direct}$} \\
        \midrule
        \textbf{CNN} \\
        direct & 0.332 & 0.422 & 0.389 & 0.400 & 0.335 & 0.376 (0.040) & \\
        transfer & 0.357 & 0.420 & 0.386 & 0.398 & 0.359 & 0.384 (0.027) & 2.2\% \\
        \midrule
        \textbf{V{\scriptsize 1}T} \\
        direct & 0.368 & 0.417 & 0.394 & 0.414 & 0.347 & 0.388 (0.030) & \\
        transfer & 0.384 & 0.450 & 0.414 & 0.415 & 0.385 & 0.410 (0.027) & 5.6\% \\
        \bottomrule
    \end{tabular}
    \end{sc}
    \end{small}
    \end{center}
\end{table}

\begin{table}[ht]
    \caption{\textbf{CNN vs V{\footnotesize 1}T cross-dataset generalization}. We first pre-trained the core module on a gray-scale version of \textsc{Dataset F}, then (\textsc{transfer}) froze the core and fine-tuned the readouts on \textsc{Dataset S}. $\Delta\textsc{original}$ shows the test performance drop in the cross-dataset transfer learning setting as compare (\textsc{original}) a model directly trained on \textsc{Dataset S}.} \label{table:cross-dataset-results}
    \vspace{-3mm}
    \begin{center}
    \begin{small}
    \begin{sc}
    \begin{tabular}{llllll|lr}
        \toprule
         & \multicolumn{5}{c}{Mouse} & \\
         & \multicolumn{1}{c}{A} & \multicolumn{1}{c}{B} & \multicolumn{1}{c}{C} & \multicolumn{1}{c}{D} & \multicolumn{1}{c}{E} & \multicolumn{1}{c}{avg (sd)} & \multicolumn{1}{c}{$\Delta\textsc{original}$} \\
        \midrule
        \textbf{CNN} \\
        original & 0.350 & 0.424 & 0.385 & 0.371 & 0.360 & 0.378 (0.029) & \\
        transfer & 0.314 & 0.353 & 0.337 & 0.316 & 0.327 & 0.329 (0.016) & -12.9\% \\ 
        \midrule
        \textbf{V{\scriptsize 1}T} \\
        original & 0.401 & 0.464 & 0.430 & 0.436 & 0.401 & 0.426 (0.027) & \\
        transfer & 0.327 & 0.382 & 0.347 & 0.343 & 0.328 & 0.345 (0.022) & -19.0\%\\
        \bottomrule
    \end{tabular}
    \end{sc}
    \end{small}
    \end{center}
\end{table}

\clearpage

\subsection{Artificial receptive fields} \label{appendix:artificial-receptive-fields}

\begin{figure}[ht]
    \centering{
        \phantomsubcaption\label{figure:CNN_aRFs}
        \phantomsubcaption\label{figure:ViT_aRFs}
        \phantomsubcaption\label{figure:aRF_centers}
    }
    \centering
    \includegraphics[width=0.91\linewidth]{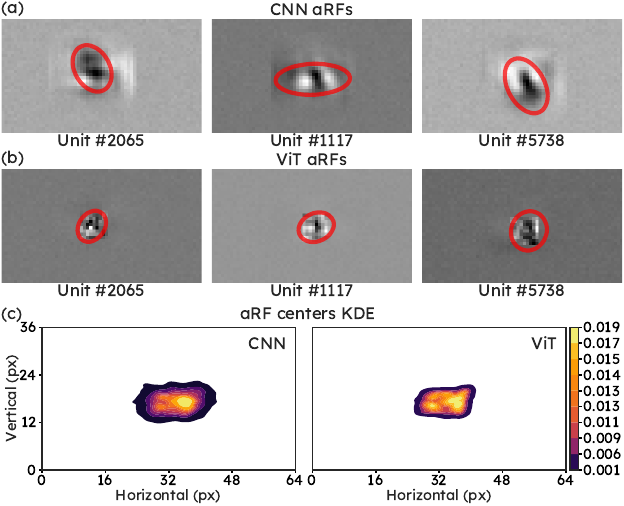}
    \vspace{-3mm}
    \caption{Estimated artificial receptive fields (aRFs) of (\subref{figure:CNN_aRFs}) CNN and (\subref{figure:ViT_aRFs}) ViT over the same set of randomly selected artificial units from \textsc{Mouse A}. The red circles (1 standard deviation ellipse) show the 2d Gaussian fit. (\subref{figure:aRF_centers}) KDE of the Gaussian centers of the two models.}
    \label{figure:aRFs}
\end{figure}

Here, we outline the procedure to estimate the artificial receptive fields (aRFs) of the CNN and ViT models (not V{\small 1}T, since there is no behavior involved) and the process to compare their spatial positions and sizes. We first present each trained model with $N=500,000$ images of white noise drawn from a uniform distribution. The aRF of unit $i$ is then computed as the summation of all noise images, weighted by the respective output: 
\begin{align}
    \text{aRF}_i = \sum_n^N \texttt{F}(x_n)_i * x_n,\quad x_n \sim \mathcal{U}^{1 \times 36 \times 64}
\end{align}
where model \texttt{F} can be either the CNN or ViT, and $\texttt{F}(x_n)_i$ denotes the response of unit $i$ given white noise image $x_n$. Figure~\ref{figure:aRFs} shows the estimated aRFs of 3 randomly selected artificial units (out of 8372 in the readout for \textsc{Mouse A}) from the two models. 

To quantify the location and size of the aRFs, we fitted a 2d Gaussian to each aRF and compared the mean and covariance of the fitted parameters. We repeated the same process for all 8372 artificial units. Concretely, we first subtracted the mean from each aRF to center the values on the baseline, then took their absolute values and fitted a 2d Gaussian using SciPy’s \href{https://docs.scipy.org/doc/scipy-1.10.1/reference/generated/scipy.optimize.curve_fit.html}{\texttt{curve\_fit()}} function. Note that not all aRFs have good fit. We thus dropped the bottom 5\% of the fitted results. Figure~\ref{figure:aRF_centers} shows the KDE plot of the fitted Gaussian means from the aRFs of the CNN and ViT. The vast majority of the aRFs are centered with respect to the image, aligning with our expectations from the attention rollout maps (see Section~\ref{results}). Figure~\ref{figure:aRF_sigma_distributions} compares the standard deviations in horizontal and vertical direction of the fitted Gaussian. 
\vspace{-3mm}

\clearpage

\subsection{Attention rollout maps} \label{appendix:attention-rollout-maps}

\begin{figure}[ht]
    \centering
    \begin{subfigure}{1\textwidth}
        \centering
        \includegraphics[width=1\textwidth]{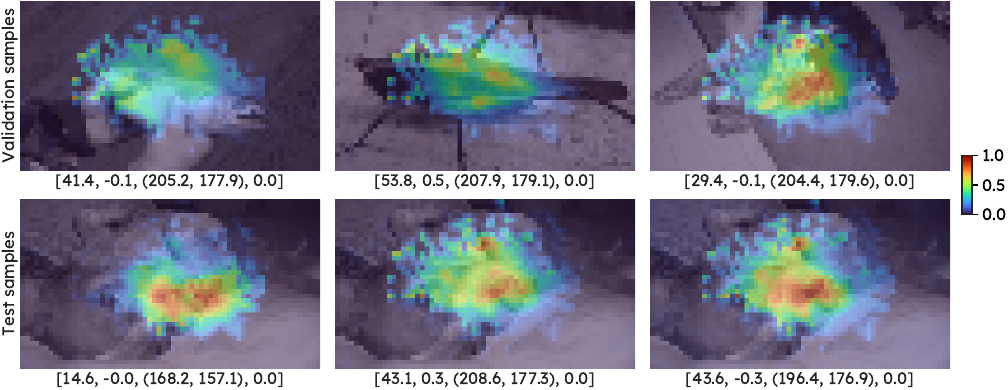}
        \caption{\label{figure:attention-rollout-mouseA} \textsc{Mouse A}} 
    \end{subfigure} \\[2mm]
    \begin{subfigure}{1\textwidth}
        \centering
        \includegraphics[width=1\textwidth]{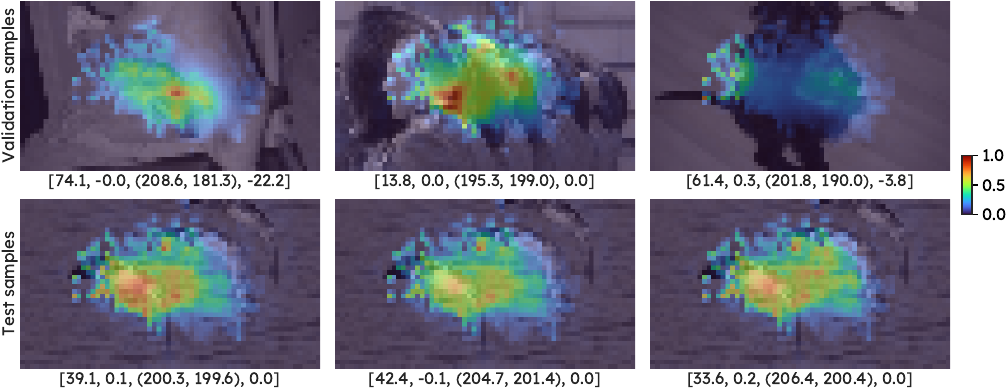}
        \caption{\label{figure:attention-rollout-mouseB} \textsc{Mouse B}} 
    \end{subfigure} \\
    \caption{V{\footnotesize 1}T attention visualization on validation and test samples of \textsc{Mouse A} to \textsc{E} from \textsc{Dataset S}. As the computer monitor was positioned such that the visual stimuli were presented to the center of the receptive field of the recorded neurons (see \textsc{Dataset S} discussion in Section~\ref{neural-data}), we expected regions in the center of the image to correlate the most with the neural responses, indicating that the core module learned to assign higher attention weights toward those regions. Note that the core module is shared among all mice. For this reason, we also expected similar patterns across animals. We observed small variations in the attention maps in the test set, where the image is the same and behavioral variables vary, suggesting the core module learned to adjust its attention based on the internal brain state. To quantify this result, we further showed that there are moderate correlations between the center of mass of the attention maps and the pupil center, see discussion in Section~\ref{results}. Each attention map was normalized to $[0, 1]$, and the behavioral variables of the corresponding trial are shown below the image in the format of [pupil dilation, dilation derivative, pupil center ($x$, $y$), speed]. The Figure continues to the next page.}
    \label{figure:attention-rollout-all}
\end{figure}
\begin{figure} \ContinuedFloat
    \begin{subfigure}{1\textwidth}
        \centering
        \includegraphics[width=1\textwidth]{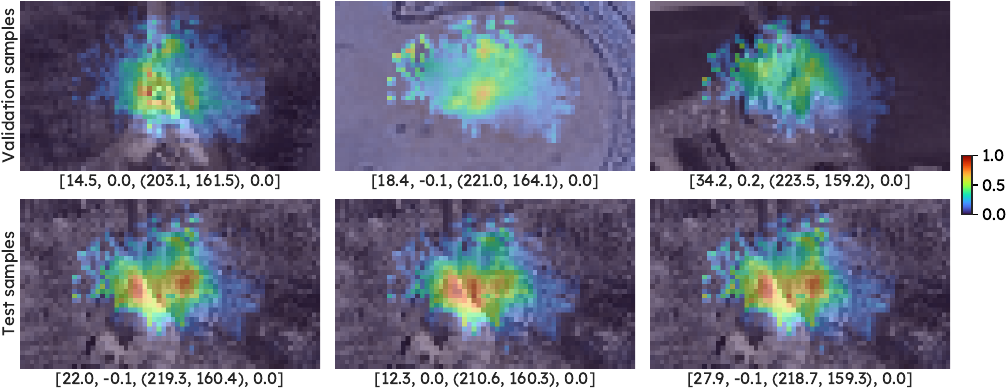}
        \caption{\label{figure:attention-rollout-mouseC} \textsc{Mouse C}} 
    \end{subfigure} \\[2mm]
    \begin{subfigure}{1\textwidth}
        \centering
        \includegraphics[width=1\textwidth]{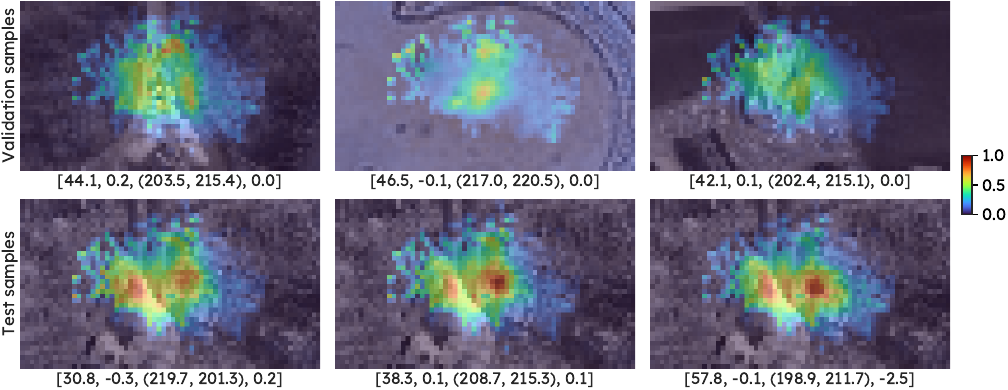}
        \caption{\label{figure:attention-rollout-mouseD} \textsc{Mouse D}} 
    \end{subfigure} \\[2mm]
    \begin{subfigure}{1\textwidth}
        \centering
        \includegraphics[width=1\textwidth]{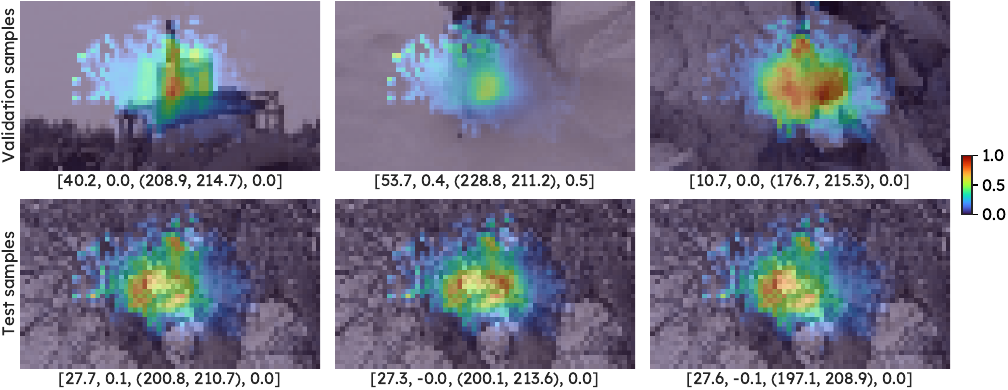}
        \caption{\label{figure:attention-rollout-mouseE} \textsc{Mouse E}} 
    \end{subfigure}
\end{figure}

\clearpage

\subsection{Behaviors and predictive performance} \label{appendix:behaviors-and-predictive-performance}

\begin{figure}[ht]
    \centering
    \includegraphics[width=0.8\linewidth]{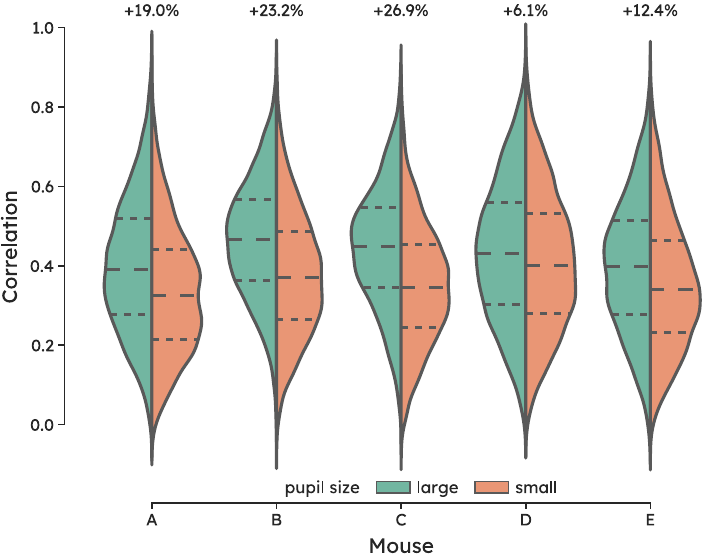}
    \caption{Predictive performance w.r.t. pupil dilation in \textsc{Dataset S}. Previous work has shown that pupil dilation is an indication of arousal, i.e. stronger (or weaker) neural responses with respect to the visual stimulus~\citep{reimer2016pupil, larsen2018neuromodulatory}. We thus expected a similar tendency could also be observed with our model. Here, we divided the test set into 3 subsets based on pupil dilation. We then compared the predictive performance of the model in the (large) larger third subset against the (small) smaller third subset. We observed that trials with larger pupil sizes are better predicted, with an average difference of $+17.5\%$ across animals. The dashed lines indicate the quartiles of the distributions and the percentage above each violin plot shows the relative prediction improvement of the larger set against the smaller set.}
    \label{figure:pupil-dilation}
\end{figure}

\clearpage

\subsection{Readout position and retinotopy} \label{appendix:readout-position}

\begin{figure}[ht]
    \centering
    \includegraphics[width=1\textwidth]{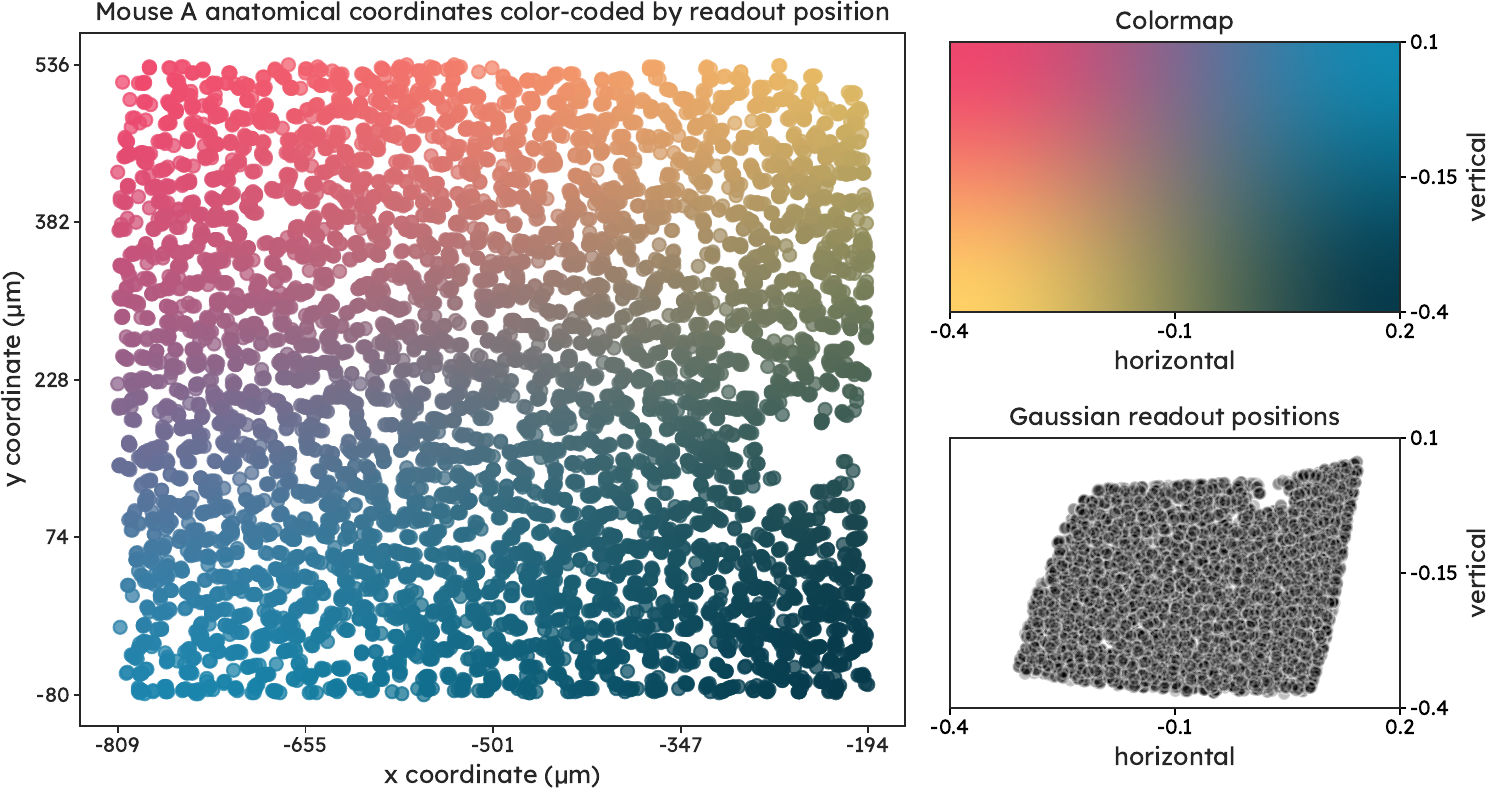}
    \caption{The learned readout position with respect to neuron anatomical coordinates in \textsc{Mouse A}. The position network in the Gaussian readout (see Section~\ref{previous-work}) learns the mapping between the latent visual representation (i.e. output of the core, bottom right panel) and the 2d anatomical location of each neuron (left panel). \citet{lurz2021generalization} and \citet{willeke2022sensorium} demonstrated that a smooth mapping can be obtained when color-coding each neuron by its corresponding readout position unit. This aligned with our expectation that neurons that are close in space should have a similar receptive field~\citep{garrett2014topography}. Here, we showed that, despite the substantial architectural change, a similar mapping can also be obtained with the V{\footnotesize 1}T core. The code to generate this plot was written by \citet{willeke2022sensorium} and is available at \href{https://github.com/sinzlab/sensorium/blob/e5017df2ff89c60a4d0a7687c4bde67774de346b/notebooks/model_tutorial/2_model_evaluation_and_inspection.ipynb}{github.com/sinzlab/sensorium}.}
    \label{figure:readout-position}
\end{figure}

\end{document}